\pdfoutput=1

\documentclass[11pt]{article}

\usepackage[]{emnlp2021}

\usepackage{times}
\usepackage{latexsym}

\usepackage[T1]{fontenc}

\usepackage[utf8]{inputenc}

\usepackage{microtype}

\usepackage{todonotes}
\usepackage{url}
\usepackage{color}
\usepackage{makecell}
\usepackage{bbm}
\usepackage{amsmath}
\usepackage{arydshln}
\usepackage{subcaption}
\usepackage{caption}
\usepackage{multirow}
\usepackage{graphicx}
\title{\textsc{BERT}, \textsc{mBERT}, or \textsc{BiBERT}? A Study on Contextualized Embeddings for Neural Machine Translation }

\author{Haoran Xu, Benjamin Van Durme,  Kenton Murray\\[1em]
Johns Hopkins University\\
\texttt{\{hxu64,vandurme,kenton\}@jhu.edu}\\[1em]
}
\date{}

\begin{document}
\maketitle
\begin{abstract}
The success of bidirectional encoders using masked language models, such as  \textsc{BERT}, on numerous natural language processing tasks has prompted researchers to attempt to incorporate these pre-trained models into neural machine translation (NMT) systems. However, proposed methods for incorporating pre-trained models are non-trivial and mainly focus on BERT, which lacks a comparison of the impact that other pre-trained models may have on translation performance. In this paper, we demonstrate that simply using the output (contextualized embeddings) of a tailored and suitable bilingual pre-trained language model (dubbed \textsc{BiBERT}) as the input of the NMT encoder achieves state-of-the-art translation performance. Moreover, we also propose a \textit{stochastic layer selection} approach and a concept of \textit{dual-directional translation model} to ensure the sufficient utilization of contextualized embeddings. In the case of without using back translation, our best models achieve BLEU scores of \textbf{30.45} for En$\rightarrow$De and \textbf{38.61} for De$\rightarrow$En on the IWSLT'14 dataset, and \textbf{31.26} for En$\rightarrow$De and \textbf{34.94} for De$\rightarrow$En on the WMT'14 dataset,  which exceeds all published numbers\footnote{Code is available at: \url{https://github.com/fe1ixxu/BiBERT}.}\footnote{Our BiBERT is released at: \url{https://huggingface.co/jhu-clsp/bibert-ende}.}. 
\end{abstract}

\section{Introduction}
Pre-trained language models (LMs), trained on a large-scale unlabeled data to capture rich representations of the input, such as \textsc{ELMo} \citep{peters-etal-2018-deep}, \textsc{BERT} \citep{devlin-etal-2019-bert}, \textsc{XLNet} \citep{NEURIPS2019_dc6a7e65} and \textsc{XLM} \citep{NEURIPS2019_c04c19c2} have increasingly attracted attention in various NLP tasks. Either utilizing context-aware representations of input tokens \citep{peters-etal-2018-deep} or fine-tuning the pre-trained parameters \citep{devlin-etal-2019-bert} both lead to significant improvement for downstream tasks. 

\begin{figure}[ht]
    \centering
    \includegraphics[width=0.48\textwidth]{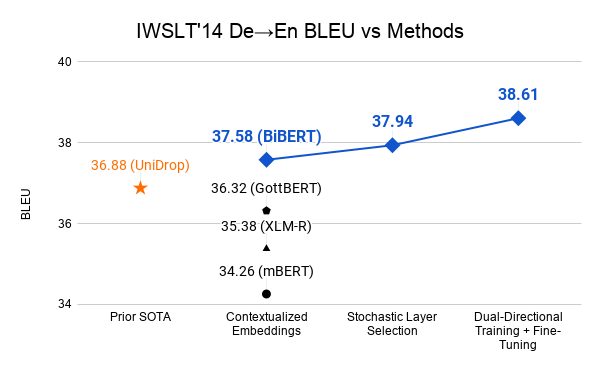}
    \caption{The overview of methods: a series of additive improvements to the use of contextualized embeddings on IWSLT'14 dataset. Experimenting over various pre-trained language models, we show that our $\textsc{BiBERT}$, a bilingual English-German language model, vastly outperforms all other methods (Section \ref{sec:contextualized_embedding}). Adding \emph{stochastic layer selection} to $\textsc{BiBERT}$ improves performance (Section \ref{sec:layer-dropout}). Finally, innovative \emph{dual-directional training} and fine-tuning with the previous two methods yield around 2 BLEU point gains over the previous state-of-the-art result \citep{wu-etal-2021-unidrop} (Section \ref{sec:bidirection-model}).}
    \label{fig:intro}
\end{figure}

Inspired by the superior performance of \textsc{BERT} on many other tasks, researchers have investigated leveraging using this pre-trained masked language model to enhance translation models, e.g., initializing the parameters of the model's encoder with \textsc{BERT} parameters \citep{rothe-etal-2020-leveraging}, and incorporating the output of \textsc{BERT} to each layer of the encoder \citep{Zhu2020Incorporating, weng2020acquiring}. In this paper, we demonstrate simply using the output of a pre-trained language model as the input of NMT systems can achieve state-of-the-art results on IWLST'14 \citep{cettolo2014report} and WMT'14 \citep{bojar2014findings} English$\leftrightarrow$German (En$\leftrightarrow$De) translation tasks in the case of without using back translation \citep{sennrich2016improving,edunov2018understanding}\footnote{Though we use a language model that has been trained with additional monolingual data, we only use the provided bitexts during machine translation training.}. After conducting a thorough evaluation of numerous pre-trained language models, we demonstrate that specialized bilingual models perform the best. We then introduce two further refinements,  \emph{stochastic layer selection} and \emph{dual-directional training} that yield further improvements. The overview of methods are shown in Figure \ref{fig:intro}. Overall, our best systems beat published state-of-the-art BLEU scores by around 2 points.

Our main contributions are listed as follows:
\begin{itemize}
  \itemsep0em 
  \item We release our English-Germean bilingual pre-trained language model, \textsc{BiBERT}, and demonstrate that it outperforms both monolingual and multi-lingual language models for machine translation (Section \ref{sec:contextualized_embedding}).
  \item Expanding upon our bilingual language model results, we introduce \textbf{stochastic layer selection} which incorporates information from more layers in the pre-trained language model to improve machine translation (Section \ref{sec:layer-dropout}).
  \item We introduce \textbf{dual-directional translation models} which leverages the inherent bilingual nature of \textsc{BiBERT} with mixed domain training and fine-tuning. When combined with stochastic layer selection, it achieves state-of-the-art performance, i.e., \textbf{30.45} for En$\rightarrow$De and \textbf{38.61} for De$\rightarrow$En on the IWSLT'14 dataset, and \textbf{31.26} for En$\rightarrow$De 
  on the WMT'14 dataset
  (Section \ref{sec:bidirection-model}).
\end{itemize}

\section{Contextualized Embeddings for NMT }
\label{sec:contextualized_embedding}
\subsection{Method}
In this section, we focus on investigating the effectiveness of using the output (contextualized embeddings) of the last layer of pre-trained language models on building NMT models. Our basic NMT models are six-layer transformer translation models, though it is model agnostic assuming there are encoder embeddings \citep{NIPS2017_3f5ee243}. Specifically, our method relies on extracting contextualized embeddings of source sentences from the final layer of a frozen pre-trained language model and feeding them to the embedding layer of the NMT encoder. Rather than randomly initializing the source embedding layer, we use the output of these pre-trained models and do not allow these parameters to update during training. To allow for a deep analysis, we concentrate on one language pair, English$\leftrightarrow$German (En$\leftrightarrow$De). In the following subsections, we first explore how much translation performance can be improved by simply using contextualized embeddings, and then explore the internal factors of various pre-trained language models that may affect NMT models. We then  introduce our bilingual pre-trained language model and demonstrate that using its contextualized embeddings achieves state-of-the-art results.

\subsection{Existing Pre-Trained Models}
We first describe four influential pre-trained models that we incorporate into NMT -- two monolingual and two multilingual models.

\paragraph{\textsc{RoBERTa}}
An optimized version of BERT which is trained on a larger dataset, with a dynamic masked language model training regiment that also removes the next sentence prediction \citep{liu2019roberta}. This model matches or exceeds the performance of BERT on multiple NLP tasks.

\paragraph{GottBERT} A state-of-the-art pure German Roberta model  \citep{scheible2020gottbert} trained on 145G German text data portion of OSCAR \citep{ortiz-suarez-etal-2020-monolingual}, a huge multilingual corpus extracted from Common Crawl. This has been shown to outperform the other two existing German monolingual models (i.e., German BERT\footnote{\url{https://deepset.ai/german-bert}} from \texttt{deepset} and \texttt{dbmz} BERT\footnote{\url{https://github.com/dbmdz/berts}}) on NER and text classification tasks.

\paragraph{\textsc{mBERT} (cased)}
A multilingual \textsc{BERT} \citep{devlin-etal-2019-bert} pre-trained on 104 highest-resource languages in Wikipedia.


\paragraph{\textsc{XLM-R} (base)} A transformer-based \cite{NIPS2017_3f5ee243} masked language model trained on 100 languages, using more than two terabytes of filtered CommonCrawl data, which outperforms \textsc{mBERT} on a variety of cross-lingual benchmarks \citep{conneau2020unsupervised}.

\subsection{How Do Pre-Trained LMs Affect NMT?}
\label{sec:affect}
First we investigate how contextualized embeddings of aforementioned pre-trained language models help NMT models, and explore possible positive and negative factors that may affect NMT models.
\paragraph{Dataset}
We initially consider a low-resource scenario and then show further experiments in a high-resource scenario in Section \ref{sec:wmt}. We conduct experiments on the IWSLT'14 English-German dataset, which has 160K parallel bilingual sentence pairs. 

\paragraph{Settings}
Our model configuration is \texttt{transformer\_iwslt\_de\_en}, a six-layer transformer architecture \citep{NIPS2017_3f5ee243}, with FFN dimension size 1024 and 4 attention heads. We use an embedding dimension of 768 to match the dimension of pre-trained language models. For a consistent comparison with previous works, the evaluation metric is the commonly used tokenized BLEU \citep{papineni2002bleu} score calculated with the \texttt{multi-bleu.perl} script. More training details are described in Appendix \ref{app:training}.

 
\begin{table}[ht]
\centering
\resizebox{1\linewidth}{!}{
\begin{tabular}{lc|cc}
\hline
\multicolumn{2}{c|}{Methods}            & En$\rightarrow$De & De$\rightarrow$En \\ \hline
\multirow{2}{*}{\textsc{RoBERTa}}  & \textit{random}      & 27.3                & -                   \\
                          & \textit{pre-trained} & 28.74($+$1.44)        & -                   \\ \hline
\multirow{2}{*}{\textsc{GottBERT}} & \textit{random}      & -                   & 33.56               \\
                          & \textit{pre-trained} & -                   & 36.32($+$2.76)        \\ \hline
\multirow{2}{*}{\textsc{mBERT}}    & \textit{random}      & 27.80               & 34.01               \\
                          & \textit{pre-trained} & \color{gray} 27.37($-$0.43)        & \color{gray} 34.26($+$0.25)        \\ \hline
\multirow{2}{*}{\textsc{XLM-R}}    & \textit{random}      & 27.87               & 33.67               \\
                          & \textit{pre-trained} & \color{gray} 27.85($-$0.02)        & 35.38 ($+$1.71)       \\ \hline
\multirow{2}{*}{\textsc{BiBERT}}   & \textit{random}      & 27.53               & 33.52               \\
                          & \textit{pre-trained} & \bf 29.65($+$2.12)        & \bf 37.58($+$4.06)        \\ \hline
\end{tabular}
}
\caption{IWSLT'14 En$\leftrightarrow$De BLEU scores utilizing contextualized embeddings from various pre-trained language models. \textit{random} represents the embedding layer of the NMT encoder that is randomly initialized but uses the same vocabulary of the assigned pre-trained language model. \textit{pre-trained} means the embedding layer of the NMT encoder use the output of the assigned frozen pre-trained language model during MT training. Numbers in the bracket show the increment/deduction compared with the corresponding model compared to randomly initialized embeddings.  }
\label{tab:iwslt14-embed}
\end{table}

\paragraph{Observations} 
The main IWSLT'14 results are shown in Table \ref{tab:iwslt14-embed}. We first conduct experiments with randomly initialized embeddings to obtain baselines. Feeding the output of a pre-trained language model into an NMT model necessitates that the vocabulary of the encoder should be the same as the one used for the language model. To ensure that improvements are not the result of choosing a better vocabulary, we train randomly initialized baseline systems using identical vocabularies for each encoder. For these experiments, the decoder's vocabulary size is fixed to 8K in order to make fair comparisons. We investigate decoder vocabulary size selection in more detail in Section \ref{subsec:vocab}. When the embedding layer of the MT encoder is randomly initialized, as opposed to using the pre-trained language model, we observe similar BLEU scores for all baselines from English-to-German (around 27.6) and German-to-English (around 33.7). By replacing the embedding layer with contextualized embeddings, \textsc{GottBERT} boosts the BLEU scores of De$\rightarrow$En from 33.56 to 36.32, and \textsc{RoBERTa} strengthens the En$\rightarrow$De translation from 27.3 to 28.74. However, the \textsc{mBERT} and \textsc{XLM-R} only provide modest improvement in De$\rightarrow$En translation and even degenerate the performance of En$\rightarrow$De translation.

\paragraph{Curse of Multilinguality}
We first note the deterioration caused by \textsc{mBERT} and \textsc{XLM-R} on En$\rightarrow$De over the randomly initialized baselines, as well as the comparatively small gains versus the monolingual models of De$\rightarrow$En. We hypothesize that contextualized embeddings from \textsc{mBERT} and \textsc{XLM-R} are hurt by the \textit{curse of multilinguality} \citep{conneau2020unsupervised}, i.e., low-resource language performance can be improved by adding higher-resource languages during pre-training, but unfortunately high-resource performance suffers and degrades. \textsc{mBERT} and \textsc{XLM-R} are trained on 100 and 104 languages respectively and the \textit{curse of multilinguality} may lead to model capacity issues that degenerate the contextualized embeddings of high-resource languages such as English and German. We attribute the slightly higher improvements of \textsc{XLM-R} over \textsc{mBERT} to the larger amounts of data used in pre-training. The large monolingual models, \textsc{RoBERTa} and \textsc{GottBERT} significantly beat a randomized baseline, but also significantly beat the multilingual models. 
Note that even though \textsc{XLM-R} has 55.6B English tokens used for pre-training, it still helps less than \textsc{RoBERTa} using around 28B English tokens, which is possibly due to interference and constrained capacity \citep{arivazhagan2019massively, johnson2017google, tan2019multilingual}. Therefore, a suitable pre-trained language model for NMT intuitively should be trained on a large amount of data, but with special care to avoid using too many languages during pre-training. 


\subsection{Customized Pre-Trained LM}
Pre-trained monolingual language models can improve performance of machine translation systems, yet 
machine translation is inherently a bilingual task. 
We hypothesize that a pre-trained language model can further improve the translation performance if \textbf{its training data is composed of a mixture of texts in both source and target languages}. In other words, we expect the source and target language data to enrich the contextualized information for each other to better facilitate translation for both directions (En$\leftrightarrow$De). Therefore, we propose our bilingual pre-trained language models, dubbed \textbf{\textsc{BiBERT}}. 

Our $\textsc{BiBERT}_\textsc{en-de}$ 
is based on the RoBERTa architecture \citep{liu2019roberta} and implemented using the \texttt{fairseq} framework \citep{ott-etal-2019-fairseq}. In order to make a direct comparison, $\textsc{BiBERT}_\textsc{en-de}$ is trained on the same German texts as \textsc{GottBERT} -- just with an additional 146GB of English texts. These are a subset of the English portion in OSCAR -- the same dataset the German texts come from. We combine English and German data and shuffle them before training. We train the model using the same number of update steps on German texts as \textsc{GottBERT}\footnote{Note that \textsc{GottBERT} uses 100K update steps, but our training data is roughly double that due to the extra English data, so we adopt 200K update steps.}. We train a unified 52K vocabulary using the WordPiece tokenizer \citep{wu2016google},  with 67GB English and 67GB German texts which are randomly sampled from the training set. $\textsc{BiBERT}_\textsc{en-de}$ is trained on TPU v3-8 for four weeks. More details about optimization for $\textsc{BiBERT}_\textsc{en-de}$ are described in Appendix \ref{app:optimization}.

\begin{figure}[ht]
     \centering
     \begin{subfigure}[b]{0.48\textwidth}
         \centering
         \includegraphics[width=\textwidth]{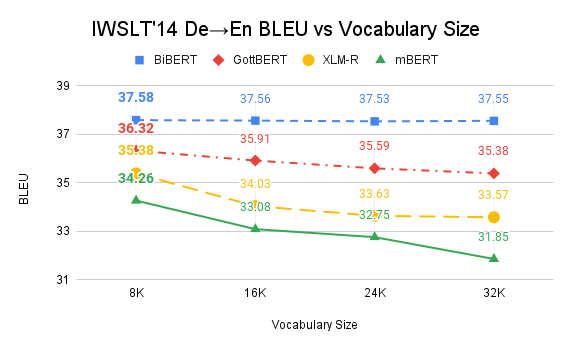}
         \caption{De$\rightarrow$En BLEU on IWSLT'14 test set}
         \label{fig:iwslt14-de-en}
     \end{subfigure}
     \hfill
     \begin{subfigure}[b]{0.48\textwidth}
         \centering
         \includegraphics[width=\textwidth]{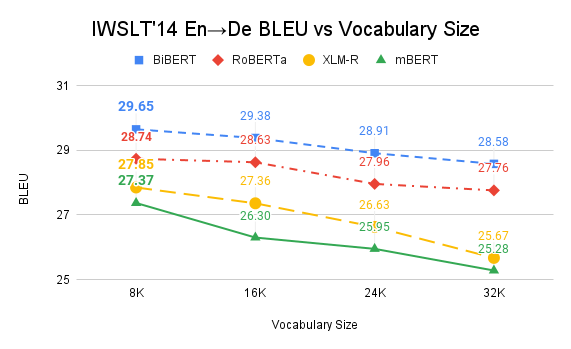}
         \caption{En$\rightarrow$De BLEU on IWSLT'14 test set}
         \label{fig:iwslt14-en-de}
     \end{subfigure}
     \caption{En$\leftrightarrow$De BLEU as a function of vocabulary size with various pre-trained language models on IWSLT'14 test set. Models obtain highest scores with 8K vocabulary size.}
     \label{fig:iwslt14}
\end{figure}


\subsection{Vocabulary Size Selection}
\label{subsec:vocab}
The vocabulary is fixed for the encoder but still indeterminate for the decoder. In a low-resource machine translation setting, performance is highly sensitive to decoder vocabulary size selection. 
\citet{gowda-may-2020-finding} demonstrated that 
a decoder vocabulary using 8K BPE operations performed best across a large grid search. 
To ensure that 8K vocabulary size is also a suitable choice for the IWSLT'14 (160K parallel sentences) dataset when combined with our method, we search over four candidate decoder vocabulary sizes (8K, 16K, 24K, and 32K) for all aforementioned pre-trained language models. As shown in Figure \ref{fig:iwslt14}, 8K yields the highest BLEU score for all of our NMT models for De$\leftrightarrow$En. Thus we select 8K as the vocabulary size of the decoder and use this for all subsequent experiments on IWSLT'14 unless otherwise noted. Interestingly, we also notice that the performance of the translation model with $\textsc{BiBERT}_\textsc{en-de}$ is robust for De$\rightarrow$En, and basically unaffected by the vocabulary size. 

\subsection{\textsc{BiBERT} Performance}
\paragraph{$\textsc{BiBERT}_\textsc{en-de}$ results} 
As indicated in the last row of Table \ref{tab:iwslt14-embed}, our bilingual model help the transformer model achieve \textbf{29.65} score for En$\rightarrow$De --- a gain of \textbf{2.12} over the baseline. For De$\rightarrow$En, our model achieves \textbf{37.58} score with a gain of \textbf{4.06}.  Recall that $\textsc{BiBERT}_\textsc{en-de}$ uses the same settings as \textsc{GottBERT} --- the only difference is the addition of extra English training data from the OSCAR corpus --- yet
$\textsc{BiBERT}_\textsc{en-de}$ yields an additional 1.30 BLEU point improvement over \textsc{GottBERT}.

\paragraph{Analysis}
Based on the superior performance of $\textsc{BiBERT}_\textsc{en-de}$, we hypothesize that contextualized embeddings output from $\textsc{BiBERT}_\textsc{en-de}$ contain richer German information than \textsc{GottBERT} and better assist the model in translation by learning extra English data. Furthermore, we theorize training on German texts also enhances the quality of English contextualized embeddings --- note that even though \textsc{RoBERTa} and $\textsc{BiBERT}_\textsc{en-de}$ are not directly comparable due to different English pre-training data, $\textsc{BiBERT}_\textsc{en-de}$ still had a 0.68 BLEU point improvement over \textsc{ROBERTa} even while using less English training data. Some other explanations for the superior performance of $\textsc{BiBERT}_\textsc{en-de}$ are 1) it learns the aligned embeddings for the tokens with similar meanings across two languages. Hence, the source embeddings can offer the encoder a hint of aligned target embeddings to help translation. 2) Embeddings of overlapping En-De sub-word units\footnote{Such as \texttt{\#\#n}, which uses shared En-De information.} fed to NMT encoders may facilitate translation by bilingual information.

\begin{table}[ht]
\begin{small}
\centering
\begin{tabular}{l|cc}

\hline
Algorithms & De $\rightarrow$ En \\
\hline
Adversarial MLE \citep{WangG019} & 35.18 \\
DynamicConv\citep{wu2018pay} & 35.20 \\
Macaron Net \citep{lu*2020understanding} & 35.40\\
BERT-Fuse \citep{Zhu2020Incorporating} & 36.11 \\
MAT \citep{fan2020multi} & 36.22  \\
Mixed Representations \citep{wu2020sequence}  & 36.41 \\
UniDrop \citep{wu-etal-2021-unidrop}  & 36.88 \\
\hline
Ours, \textsc{GottBERT} & 36.32 \\
Ours, \textsc{BiBERT} & \bf 37.58 \\

\hline

\end{tabular}
\caption{Comparison of our work and most recent existing methods on IWSLT'14 De$\rightarrow$En.}
\label{tab:iwslt14-previous}
\end{small}
\end{table}

\begin{figure*}[ht]
    \centering
    \includegraphics[width=11.5cm]{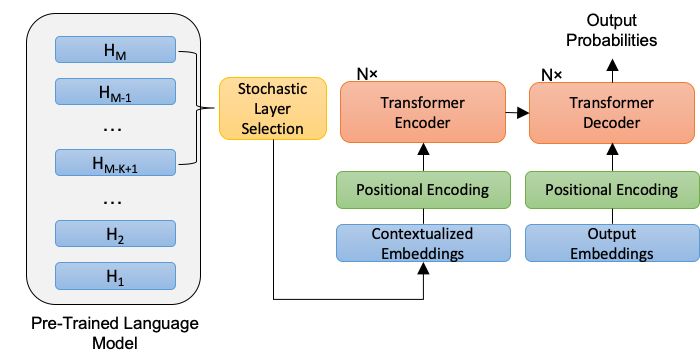}
    \caption{The overall framework of \textit{stochastic layer selection} method. Top $K$ layers of the  pre-trained language model are considered and fed to the NMT encoder. }
    \label{fig:dropout}
\end{figure*}

\subsection{Comparison with Existing Work}
Table \ref{tab:iwslt14-previous} shows a comparison of our work with the recent literature on IWSLT'14 German to English translation. These works propose improvements to transformer models in different aspects, e.g., incorporating BERT into every layer of encoders and decoders with additional multi-head attentions \citep{Zhu2020Incorporating}, multi-branch encoders \citep{fan2020multi}, mixed representations from different tokenizers \citep{wu2020sequence} and uniting different dropout techniques into NMT models \citep{wu-etal-2021-unidrop}. Our straightforward method of simply using the final layer of $\textsc{BiBERT}_\textsc{en-de}$ outperforms all of them. Furthermore, even the model that only uses the monolingual \textsc{GottBERT} achieves a competitive result (36.32) compared with the previous state-of-the-art approach (36.88). Our method is easy to implement, so it can be used in conjunction with other methods in the literature.

\subsection{Time Costs}
Leveraging an external pre-trained language model leads to higher computational complexity. Our approach takes approximately 20\% additional time during training and 13\% extra time during inference. Considering the significant BLEU gains, we argue that they justify the higher time costs. 

\section{Layer Selection}
\label{sec:layer-dropout}
\citet{jawahar-etal-2019-bert} demonstrates that different layers of BERT capture differing linguistic information in a rich, hierarchical structure that mimics classical, compositional tree-like structures. 
Information in the lower layer (e.g., phrase-level information) gets gradually diluted in higher layers. Thus, to potentially leverage more information encapsulated in the pre-trained language models, we are also interested in exploring how other layers  of contextualized embeddings can improve NMT models --- rather than simply using the last layer. 

We denote $\mathcal{X}$ as the collection of source language sentences. For each source sentence $x\in\mathcal{X}$, let $H_B^i(x)$ denote the contextualized embeddings of $x$ obtained from the $i_{th}$ layer of the pre-trained language model. In our settings, we consider top $K$ layers of the pre-trained language model, i.e., we consider $H_B^i(x)$ $\forall i\in [M-K+1, M]$, where $K$ is a hyperparameter, and $M$ is the total number of layers of the pre-trained language model. 


\subsection{Stochastic Layer Selection}
During training of deep neural networks,  various methods of stochastically freezing groups of parameters in a model for individual training examples have been shown to improve performance. For instance, dropout \citep{srivastava2014dropout} samples parameters from a Bernoulli distribution to not update, and drop-net \citep{Zhu2020Incorporating} and drop-branch \cite{fan2020multi} randomly active a candidate net and freeze the others in a uniform distribution. We propose \textit{stochastic layer selection}, a novel approach to encapsulate more features and information from more layers of the pre-trained language models. 
Specifically, for each batch, we randomly pick the output from one layer rather than all of them as the input for the NMT encoder (Figure \ref{fig:dropout}). We denote the input embeddings of sentence $x$ to the NMT encoder as $H_E(x)$, which is defined in the following way during training:

\begin{equation}
    H_E(x) = \sum_{i=1}^K \mathbbm{1}(\frac{i-1}{K}<p\leq\frac{i}{K})H_B^{M-i+1}(x)
    \label{eq:drop-layer}
\end{equation}
where $\mathbbm{1}(\cdot)$ is the indicator function and $p$ is a random variable which is uniformly sampled from [0,1]. In the inference step, the output is the expectation of outputs of all layers used for training, i.e., $\mathbbm{E}_{p\sim\text{uniform}[0,1]}[H_E(x)]$, which leads to the modification of Equation \ref{eq:drop-layer}:

\begin{equation}
    H_E(x) = \frac{1}{K} \sum_{i=1}^K H_B^{M-i+1}(x).
    \label{eq:mean}
\end{equation}

\subsection{Experiments and Results}
\begin{figure}[t]
    \centering
    \includegraphics[width=7.6cm]{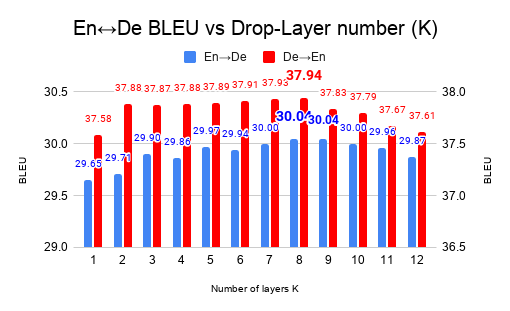}
    \caption{IWSLT'14 En$\rightarrow$De (left, blue bars) and De$\rightarrow$En (right, red bars) BLEU as a function of number of layers $K$ considered in the \textit{stochastic layer selection} module for NMT models. Note that when $K=1$, it reduces to the case of selecting the last layer. However, any value of $K>1$ selected for stochastic layer selection beats this very strong baseline with $K=8$ obtaining the highest BLEU scores in both directions.}
    \label{fig:drop}
\end{figure}

Based on the results of Table \ref{tab:iwslt14-embed}, we select the pre-trained model performing best for NMT, $\textsc{BiBERT}_\textsc{en-de}$, and use it as the basis for
all subsequent experiments. To be consistent with the results in Section \ref{sec:contextualized_embedding}, we once again use the IWSLT'14 dataset. Figure \ref{fig:drop} illustrates the impact of stochastic layer selection. We conduct experiments for En$\leftrightarrow$De with the number of layers $K$ ranging from 2 to $M$ ($M$ = 12 for $\textsc{BiBERT}_\textsc{en-de}$). Note that setting $K=1$ reduces to the case of only selecting  the last layer as in Section \ref{sec:contextualized_embedding}.

In all cases, the stochastic layer selection obtains substantial gains compared with our previous best scores in En$\rightarrow$De (29.65) and De$\rightarrow$En (37.58) in Section \ref{sec:contextualized_embedding}. In both situations of En$\rightarrow$De and De$\rightarrow$En, the translation model gets the highest score (\textbf{37.94} for De$\rightarrow$En and \textbf{30.04} for En$\rightarrow$De) when stochastic layer selection uses 8 layers. 

\section{One Model, Dual-Directional Translations}
\label{sec:bidirection-model}
In this section, different from ordinary one-way translation models, we introduce our dual-directional translation models, i.e., a model can translate both En$\rightarrow$De and De$\rightarrow$En. The model architecture is the same as the one in Section \ref{sec:layer-dropout}.

One of the biggest advantages of the shared English-German vocabulary of $\textsc{BiBERT}_\textsc{en-de}$ is that our encoder has the capability of receiving contextualized embeddings of both source and target tokens. During the training step, we feed source sentences to the model and expect the generation of a target translation, yet also, inversely, feed target sentences and expect translations in the source language. The motivation behind the dual-directional translation model is that we expect the contextualized representations of source and target sentences could enhance each other to build a better encoder for the translation model. From the aspect of data augmentation, the target sentences play a role in augmented data in the task of translating from the source language to the target language, vice versa. With the method of swapping source and target sentences once as an additional dataset, our experiments show superior performance for both directional translations. Two advantages of this method are 1) obtaining improvement without extra bitexts, and 2) only slight modification for data preprocessing and no changes for the model architecture. 
\begin{figure}[t]
    \centering
    \includegraphics[width=5.5cm]{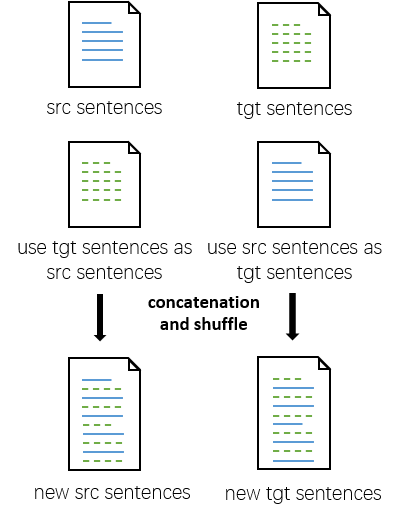}
    \caption{Workflow of data preprocessing. We swap source and target sentences, and concatenate swapped sentence pairs and original sentene pairs. Finally, we shuffle the concatenated data for dual-directional translation model training.  }
    \label{fig:bi}
\end{figure}

\subsection{Dataset Preprocessing}
For consistent comparisons, the dataset is still IWSLT'14 En$\rightarrow$De. The details of data preprocessing for the dual-directional translation model are illustrated in Figure \ref{fig:bi}. Using only the same exact parallel sentences in our bitext for training, we simply leverage the dataset in reverse, by swapping our original target sentences to use as new source sentences and original source sentences as new target sentences. We then concatenate and jointly shuffle the original and new data to acquire our mixed training data. We use a joint English-German vocabulary of size 12k for the decoder. 


\subsection{Fine-tuning}
Inspired by the findings of \citet{xu-etal-2021-gradual}, where training on a mix of in- and out-of-domain of data initially, and then gradually fine-tuning until only in-domain data is used, substantially improved model performance, we treat our concatenated sentences as mixed domain data, and the source and target languages are separate language domains. Each language data can be the out-of-domain data for the other language. Following this perspective, we first train our dual-directional model on mixed data, and then fine-tune it on the source or target data to obtain one-way translation models\footnote{In the view of the same amount of training data for the source and target language, we only apply one-stage fine-tuning, which is slightly different from the multiple stages of fine-tuning in \citet{xu-etal-2021-gradual}.}. 

\subsection{Experiments and Results}
We additionally conduct one-way translation models with 12K bilingual vocabulary to have a fair baseline for dual-directional models. Overall results are shown in Table \ref{tab:iwslt14-bi}. We first discuss the models trained without stochastic layer selection. The dual-directional model substantially outperforms the one-way model by obtaining a gain of 0.52 in En$\rightarrow$De and 0.72 in De$\rightarrow$En. Moreover, fine-tuning on the in-domain data further improves BLEU from 29.89 to 30.33 in En$\rightarrow$De and from 37.97 to 38.12 in De$\rightarrow$En. Both positive results indicated by the dual-directional model and fine-tuning approach show their effectiveness in helping translation. A similar discussion holds for the models with the stochastic layer selection method. Compared with our previous models in Section \ref{sec:layer-dropout} (30.04 En$\rightarrow$De and 37.94 in De$\rightarrow$En), our best model achieves new state-of-the-art results both in En$\rightarrow$De and De$\rightarrow$En, which respectively obtain \textbf{30.45} and \textbf{38.61} BLEU. 


\begin{table}[ht]
\begin{small}
\centering
\begin{tabular}{lcc}
\hline
Methods  & En$\rightarrow$De & De$\rightarrow$En \\ 
\hline
\textit{No Stochastic Layer Selection}:  &    &    \\ 
\hline
\multicolumn{1}{l|}{One-Way (vocab size=12K)} & \multicolumn{1}{c}{29.37}   & 37.25 \\
\multicolumn{1}{l|}{Dual-Directional Training} & 29.89 & 37.97 \\
\multicolumn{1}{l|}{\quad $+$ Fine-Tuning} & \multicolumn{1}{c}{\textbf{30.33}} & \textbf{\textbf{38.12}} \\

\hline
\textit{Stochastic Layer Selection, $K=8$}:  &    &    \\ 
\hline
\multicolumn{1}{l|}{One-Way (vocab size=12K)} & \multicolumn{1}{c}{30.00}   & 37.69 \\
\multicolumn{1}{l|}{Dual-Directional Training} & 30.30 & 38.37 \\
\multicolumn{1}{l|}{\quad $+$ Fine-Tuning } & \multicolumn{1}{c}{\textbf{30.45}} & \textbf{\textbf{38.61}} \\
\hline

\end{tabular}
\caption{Comparison of dual-directional and ordinary (one-way) translation models, with and without stochastic layer selection, on IWSLT'14 En$\leftrightarrow$De. }
\label{tab:iwslt14-bi}
\end{small}
\end{table}

\begin{table*}[ht]
\begin{small}
\centering
\begin{tabular}{lcc}

\hline
Methods & En $\rightarrow$ De & De $\rightarrow$ En \\
\hline
Transformer $+$ Large Batch  \citep{ott-etal-2018-scaling}  & 29.3 & - \\
Evolved Transformer  \citep{so2019evolved} & 29.8 & - \\
\textsc{BERT} Initialization (12 layers) \citep{rothe-etal-2020-leveraging} & 30.6 &  33.6 \\
\textsc{BERT}-Fuse \citep{Zhu2020Incorporating} & 30.75 & - \\

\hline
\textsc{BiBERT} Contextualized Embeddings $+$ Stochastic Layer Selection & 30.91 & \bf 34.94 \\
\quad $+$ Dual-Directional Training & 30.31 & 34.54 \\
\quad $+$ Fine-Tuning & \bf 31.26 & 34.68 \\
\hline

\end{tabular}
\caption{WMT'14 En$\leftrightarrow$De results on \texttt{newstest2014} test set.}
\label{tab:wmt}
\end{small}
\end{table*}

\section{High-Resource Scenario}
\label{sec:wmt}

\subsection{Dataset and Training Details}
For the high-resource scenario, we evaluate models on the WMT'14 English-German dataset, which contains 4.5M parallel sentence pairs. We combine \texttt{newstest2012} and \texttt{newstest2013} as the validation set and use \texttt{newstest2014} (3003 sentence pairs) as the test set. Our model configuration is \texttt{transformer\_vaswani\_wmt\_en\_de\_big}, a `big' transformer with 4096 FFN dimension and 16 attention heads. We replace hidden size 1024 with 768 to keep the dimensions consistent with $\textsc{BiBERT}_\textsc{en-de}$. The evaluation strategy is the same as IWSLT'14 tasks.

Following the findings that En$\leftrightarrow$De translation has similar results for vocabularies ranging from 32K to 64K in high-resource scenarios (4.5M training samples) \citep{gowda-may-2020-finding}, we use a bilingual vocabulary with 52K size for the decoder, which is larger than the ones (8K and 12K) used in IWSLT experiments.

\subsection{Results}
We compare our methods with prior existing works that achieve highest scores by only using provided bi-texts in Table \ref{tab:wmt}. With $\textsc{BiBERT}_\textsc{en-de}$ contextualized embeddings and stochastic layer selection, our model achieves state-of-the-art BLEU both on En$\rightarrow$De (30.91) and De$\rightarrow$En (34.94). Interestingly, dual-directional translation training does not show the same strong effectiveness as it did in the low-resource scenario. One possible reason is that model capacity is not large enough to handle mixed domain data \citep{arivazhagan2019massively}. However, it still additively improves En$\rightarrow$De to \textbf{31.26} BLEU. It is worth mentioning that our NMT model achieves better performance with less training parameters --- the hidden size of our NMT model is 768 but 1024 for the prior existing works.

\section{Related Work}

\subsection{Pre-Trained Embeddings}
Traditional pre-trained embeddings are investigated in type level, e.g., word2vec \cite{mikolov2013efficient}, glove \citep{pennington2014glove} and fastText \citep{bojanowski-etal-2017-enriching}. \citet{peters-etal-2018-deep} moved further from this line and proposed context-aware embeddings output from pre-trained bidirectional LSTM (\textsc{ELMo}). Following the attention-based transformer module \citep{NIPS2017_3f5ee243}, the architectures of \textsc{GPT} models \citep{radford2018improving, radford2019language, NEURIPS2020_1457c0d6} and \textsc{BERT} \citep{devlin-etal-2019-bert} respectively are based on stacking deep transformer decoders and encoders and significantly boost downstream tasks. Beyond pure English models, pre-trained language models for other languages have also showed up, e.g., \textsc{CamemBERT} for French \citep{martin2020camembert} and  \textsc{AraBERT} for Arabic \citep{baly2020arabert}.
Multilingual representations, e.g. \textsc{mBERT} and \textsc{XLMs} \citep{NEURIPS2019_c04c19c2} have been shown to be effective to facilitate cross-lingual learning. \textsc{XLM-R} \citep{conneau2020unsupervised}, a model learning cross-lingual representation at scale achieved state-of-the-art results on multiple cross-lingual benchmarks. Recently, an English-Arabic bilingual BERT \citep{lan-etal-2020-empirical} outperformed \textsc{AraBERT}, \textsc{mBERT} and \textsc{XLM-R} on supervised and zero-shot transfer settings.

\subsection{MT with Context-Aware Representations}
\citet{imamura-sumita-2019-recycling} removed the NMT encoder part and directly fed the output of \textsc{BERT} to the attention mechanism in the decoder. They train the model with two optimization stages, i.e.,  only training the decoder and fine-tuning BERT. Similarly, \citet{clinchant-etal-2019-use} have incorporated \textsc{BERT} into NMT models by replacing the embedding layer with \textsc{BERT} parameters and initializing encoder with \textsc{BERT}, but they still notice that NMT model with \textsc{BERT} is not as robust as expected. \citet{rothe-etal-2020-leveraging} also leveraged pre-trained checkpoints (e.g., \textsc{BERT} and \textsc{GPT}) to initialize 12-layer NMT encoder and decoder and achieved state-of-the-art results. Interestingly, they showed that the models with decoder initialized by \textsc{GPT} fail to improve the translation performance and are even worse than the one whose decoder is randomly initialized. Similarly, \citet{ma2020xlm} initialize both transformer encoder and decoder by \textsc{XLM-R} but fine-tune it on multiple bilingual corpora to obtain a multilingual translation model. The preliminary experiments from \citet{Zhu2020Incorporating} indicate that NMT models simply fed by the output of \textsc{BERT} outperform the models initialized by \textsc{BERT} or \textsc{XLM}. However, only limited experiments and little analysis on this method has been done in their work. They mainly focused on the \textit{BERT-fuse} approach, i.e., the output of \textsc{BERT} is fed to each layer of NMT encoder and decoder with extra multi-head attentions. Instead of only using the last layer of \textsc{BERT}, \citet{weng2020acquiring} introduced layer-aware attention mechanism to capture compound contextual information from \textsc{BERT}. Moreover, they also proposed the knowledge distillation paradigm to learn pre-trained representation in the training process. On an English-Arabic translation task, \citet{yarmohammadi-etal-2021-everything} use a precursor of this method though it lacks all of the refinements described here. However, it was shown to further help in downstream cross-lingual information extraction tasks.

\section{Conclusion}
We have shown that our \textsc{BiBERT} trained on a large amount of mixed texts of the source and target languages can better help NMT models improve translation performance compared with other existing pre-trained language models and achieve state-of-the-art results by simply using the output of the last layer. Moreover, we introduce the stochastic layer selection method and demonstrated its effectiveness in improving translation performance. Finally, experiments on the dual-directional translation model illustrate that source and target data can augment each other to further boost  performance.

\section*{Acknowledgements}
We thank anonymous reviewers for their valuable comments. We thank Kelly Marchisio and Kevin Duh for their helpful suggestions. We also thank the Google TFRC program for providing free TPU access. This work was supported in part by IARPA BETTER (\#2019-19051600005). The views and conclusions contained in this work are those of the authors and should not be interpreted as necessarily representing the official policies, either expressed or implied, or endorsements of ODNI, IARPA, or the U.S. Government. The U.S. Government is authorized to reproduce and distribute reprints for governmental purposes notwithstanding any copyright annotation therein.

\bibliography{anthology,custom}
\bibliographystyle{acl_natbib}

\clearpage

\appendix
\section{Training Details}
\label{app:training}
\subsection{IWSLT'14 Training Details}
We use 4 NVIDIA 2080 Ti GPUs with 2048 tokens per GPU and accumulate the gradient 4 times. The learning rate is 0.0004. The optimizer is Adam \citep{kingma2014adam} with \texttt{inverse\_sqrt} learning rate scheduler. At inference time, we use beam search with width 4 and use a length penalty of 0.6 \citep{boulanger2013audio,wu2016google,koehn-knowles-2017-six}. 

\subsection{WMT'14 Training Details}
We use 4 NVIDIA V100 GPUs with a batch size of 4096 tokens per GPU. Following the recommendation of the training settings from \citet{ott-etal-2018-scaling}, we accumulate the gradient 32 times to simulate 128 GPU training settings. We set the initial learning rate as 0.001.

\section{Optimization for \textsc{BiBERT}}
\label{app:optimization}
We use \texttt{fairseq} to pre-train our bilingual language models on an 8-core TPU v3-8. We train the models in 200K update steps using a batch size of 8192. We use Adam optimizer \citep{kingma2014adam} with a learning rate of 4e-4, $\beta_1 = 0.9$, $\beta_2 = 0.999$, L2 weight decay of 0.01. The learning rate is warmed up over the first 20K steps to a peak value of 4e-4, from which the learning rate polynomially decayed. We apply a dropout rate of 0.1 to all layers.

\end{document}